# Anubhuti – An annotated dataset for emotional analysis of Bengali short stories

ADITYA PAL[1], BHASKAR KARN[2]
*Computer Science and Engineering*
*Birla Institute of Technology, Mesra*
Ranchi, India
[1]aditya.pal.science@gmail.com
[2]bhaskar@bitmesra.ac.in

*Abstract*— Thousands of short stories and articles are being written in many different languages all around the world today. Bengali, or Bangla, is the second highest spoken language in India after Hindi and is the national language of the country of Bangladesh. This work reports in detail the creation of Anubhuti - the first and largest text corpus for analyzing emotions expressed by writers of Bengali short stories. We explain the data collection methods, the manual annotation process and the resulting high inter-annotator agreement of the dataset due to the linguistic expertise of the annotators and the clear methodology of labelling followed. We also address some of the challenges faced in the collection of raw data and annotation process of a low resource language like Bengali. We have verified the performance of our dataset with baseline Machine Learning as well as a Deep Learning model for emotion classification and have found that these standard models have a high accuracy and relevant feature selection on Anubhuti. In addition, we also explain how this dataset can be of interest to linguists and data analysts to study the flow of emotions as expressed by writers of Bengali literature.

*Keywords*— Emotional Analysis, Bangla, Text Corpus, Annotated Dataset, Machine Learning

## I. Introduction

Emotional analysis of textual data is the process of identification, extraction and analysis of emotions or feelings as expressed by the writer of the text. Great efforts are being made to generate the most human like emotionally accurate speaking agent. Emotional analysis has gained a lot of popularity recently and is fast becoming an important area of study in computational linguistics. With advancements in the domains of speech-recognition and natural language processing, emotion detection has found a wide range of applications such as creating the most effective and human-like communication systems [1], identifying abuse and hate messages from social media [2] and a variety of studies on the psychological aspects of human behaviour.

According to Ethnologue, as of 2020, Bengali or Bangla is the seventh most popular language in the world with an estimated 265 million speakers. It is the second most popular language in India behind the national language Hindi. As mentioned by Google and KPMG in a joint report in 2017 [3], Bengali was used by over 50 million users as their primary language in India in 2016. As more and more native speakers are starting to use the internet, big technology companies are investing heavily on bringing support for regional languages to their platforms. Regional languages like Bengali have extremely low resources and datasets available on the internet. Most of the studies on sentiment analysis and emotional analysis have been performed for highly popular languages like English. However, emotion detection of regional languages like Bengali is still in a very nascent stage and a lot needs to be explored. Hence, the introduction of our dataset is relevant and well-timed.

## II. Related Works

Before discussing about the techniques and work related to emotional analysis of textual data, it is important to talk about sentiment analysis. A lot of research has been conducted on sentiment analysis in the past few years for a wide range of languages. Sentiment analysis and emotional analysis both attempt to derive feelings as expressed by the author of the text under consideration, but there are a few differences between them. Sentiment analysis, in the crude sense of the term is a technique to categorize a text element into a positive or a negative sense only. Emotional analysis on the other hand seeks to obtain not just these positive-negative polarities, rather a more diverse range of emotional tones from the text. Emotional analysis is often considered a more challenging task than sentiment analysis, especially for low-resource languages like Bengali.

Emotional analysis from textual data has mainly been attempted through four types of tasks - Emotion Polarity Classification which is similar to sentiment analysis, Emotion Classification, Emotion Detection and Emotion Cause Detection. We limit our work to the creation of a dataset for accomplishing the task of Emotion Classification only. Nasukawa and Yi have performed sentiment analysis for the English language based on blogs and web articles [4]. Recently analysing sentiments from Twitter data [5] has also received attention due to the wide range of emotions captured in user tweets which are easily accessible via public Application Programming Interfaces (APIs). Some work has been done to create datasets and perform sentiment analysis of Bangla microblog posts [6], [7]. Sumit et al have explored word embeddings for Bangla Sentiment Analysis and show that Word2vec Skip-Gram model outperforms other models [8]. However, resources for emotional analysis of Bangla text are few and far between. Tripto and Ali have explored emotions expressed by YouTube comments on Bangla videos [9]. They claim to have created a dataset with annotated emotions to train their models and derive results. Apart from this there has not much work been done to establish datasets for emotion analysis of Bengali short stories. Hence our work seeks to shed light on a low-resource language like Bengali and how a writer expresses emotions through their stories.

## III. Dataset Creation

Anubhuti was created with an intention of generating a dataset for emotional analysis of Bengali Short stories. So, we found a list of some of the most popular Bengali stories available online from several sources and a variety of authors. In all, we collected 159 unique stories spanning across 10 genres like প্রেমকাহিনী (Romance) and রহস্যগল্প (Mystery). Not only did we use the sentences from these stories, we wanted to capture the context of these texts. Hence, we decided to preserve the order of these sentences for ease of the annotators and enabling them to follow a context-based approach for annotation. We also searched for audio books on these short stories whenever possible which helped our annotators decode some of the context-based sentences in our text and provide annotations using context in case of confusion between one or more emotion types. In all, the Anubhuti dataset has been annotated by 3 native Bengali speakers having more than 15 years of formal education each in Bengali language. The annotators were given full resources to understand the corpus and they were quite happy to read and annotate the stories.

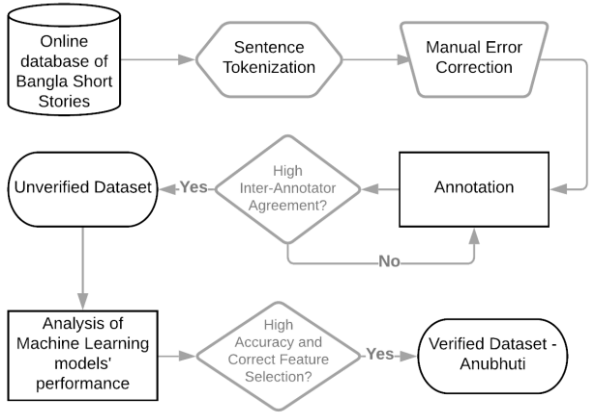

Fig. 1. Creation and Analysis of the Anubhuti dataset

The text from these 159 stories was tokenized into individual sentences using a Python Script for automation. A few noise elements had appeared during this process like incorrect tokenization at semi-colon (;) or when no Devanagari phase separator (।) was provided. The annotators were asked to first resolve these inaccuracies and provide corrected sentences from these automated elements. Once, the opinions of the annotators were gathered individually, we generated an improved set of sentences from this corpus and distributed it to the annotators to start the annotation process. The final dataset contained more than 30000 sentences. The annotation process was completed in three months and was reviewed by the authors for another month. "Fig. 1" shows the order of steps executed to create Anubhuti from start to end.

The categories to annotate the text into were evaluated based on an initial investigation on the basic emotions by Plutchik [10]. By taking a closer look at our dataset and the proportion of sentences making up all the basic emotions for a small subset of Anubhuti, we figured out that not all the basic emotions were prominent for Bengali text. We eventually came to the conclusion that four emotional categories were prevalent in the Anubhuti Corpus – Joy, Anger, Sadness, Suspense. All other statements have been grouped in a Neutral category which signifies no particular emotion category for that sentence in the dataset.

We have used Fleiss' Kappa [11] to measure the inter-annotator reliabilities for the individual emotional categories as well as for the entire Anubhuti dataset over Krippendorff's alpha [12]. "Fig. 2" summarizes the results of these inter-annotator metrics and also lists the number of sentences per category. The high inter-annotator agreement clearly shows the linguistic expertise of the 3 annotators and proves that the annotation process was conducted with a clear set of defined rules. "Fig. 3" provides a few examples of sentences along with their English translations relevant to each of these emotional categories.

TABLE I.  ANUBHUTI INTER-ANNOTATOR RELIABILITY

| Emotion Category | Sentence Count | Fleiss' Kappa |
|---|---|---|
| Joy | 5308 | 0.867 |
| Anger | 3490 | 0.821 |
| Sorrow | 3785 | 0.884 |
| Suspense | 2986 | 0.743 |
| Neutral | 16555 | 0.782 |
| Full Dataset | 32124 | 0.807 |

Fig. 2. Sentence count and inter-annotator reliability metrics per category

TABLE II.  EXAMPLES OF ANUBHUTI

| Emotion | Anubhuti Sentence | English Translation |
|---|---|---|
| Joy | মায়ের হাতের খাবার খেয়ে তার মুখে হাসি ভরে গেল। | A smile filled his face on having food cooked by his mother. |
| Anger | রাধা তার পরিবারের সকল সদস্যকে অপমান করছে। | Radha is insulting all the members of her family. |
| Sorrow | দু'চোখ দিয়ে জল গড়িয়ে পড়ে বৃন্দার। | Tears rolled out of Brinda's eyes. |
| Suspense | আমি ভয়ে আর ওখানে দাঁড়াতে পারিনি। | I could not run there out of fear. |
| Neutral | তারা রাস্তা থেকে নেমে দাঁড়ায় তাদের ধানের জমির আলের উপর। | They got down from the road and stood on the paddy field. |

Fig. 3. Examples of Anubhuti sentences with their English translations

IV. ANNOTATION CHALLENGES

Yadolahi et al [13] have already established that computational methods used in sentiment analysis tasks can be applied to emotion analysis tasks. Hence the process of annotation of a corpus for emotion analysis poses similar challenges as for sentiment analysis of English language. However, there are additional language-specific challenges that were faced during the annotation process of Bengali text. In order to fully understand the language specific challenges, we have put forth some of the common issues faced by the annotators of Anubhuti dataset.

Spelling Variations – Some Bengali words are spelt differently by different authors depending on the dialect preferred by them but in fact mean the same. Traditional Machine Learning and Deep Learning models must learn to account for these variety of spellings.

Morphological Variations – Indic languages like Hindi and Bengali are morphologically rich, which means that words can convey much more information than morphologically weaker languages like English. For example, the words আপনি, তুমি and তুই - all translate to the same meaning of 'you' in English. However, there is a significant side information provided by these words. আপনি is generally used when we address someone senior or elder while তুই is used to address a junior or younger person. তুমি is a used generally to address a peer or colleague. Such additional information needs to be captured by the Machine Learning models which makes the training of morphological rich languages like Bengali challenging, more complex and error-prone.

Order of words – A major difference between Bengali and English is the significance of word order to establish correctness of a sentence. English language is considered to have a fixed order, while Indic languages like Bengali are free order languages. A grammatically correct sentence in English has the following specific order only – Subject (S) followed by a Verb (V), which is in turn followed by Subject 2 (T). An example of this would be "Aditya drank water" which translates to "Aditya (S) খেলো (V) জল (T)" in Bengali. However, Bengali language allows other variations of this order like "Aditya জল খেলো" |STV| or "জল খেলো Aditya" |TVS| and all of these are grammatically correct. The presence of free word order in Bengali language poses challenges in contextual understanding of Machine Language models.

Apart from these language specific challenges, there were also a few context specific annotation challenges as well. Some sentences when considered in isolation may have a separate meaning than when the sentence is read in the context of the story or paragraph. An example of this was the presence of sarcasm in the text. Sarcasm based sentences are mostly preceded by an angry or a happy context making it difficult for annotators to extract the correct category of the sentence. The annotators of Anubhuti were given a clear instruction to differentiate the sarcasm based on the context.

In order to make the annotation task easier and more transparent, a few guidelines were established which helped in obtaining unbiased verdicts from each of the 3 annotators. For example, the following sentence "মোহনবাগান ইস্টবেঙ্গল কে হারালো |" translates to the football team Mohun Bagan defeating the East Bengal team. This statement may tend to be categorized as happiness by a Mohun Bagan fan while an East Bengal supporter may annotate it as sorrow. An unbiased reader should annotate it as a neutral sentence when neither supporter's perception is taken into account. We have provided a guideline to help annotators decide a category in case of confusions – Identify the reason and possible candidates of categories, listen to audio books if available to understand context, read a sentence as intended by the author going beyond the barriers of geography, religion and other differences, and if confusion persists – mark the sentence as neutral.

## V. DATASET ANALYSIS

In this section, we have provided a statistical description of the Anubhuti dataset. From "Fig. 2", we find that the dataset is imbalanced towards neutral sentences as the total number of neutral sentences exceeds the sum of other categories. Neutral sentences constitute the highest percentage (~52%) of the dataset while suspense constitutes the least amount (~9%) of the dataset. We have also provided a list of genres of the stories used for the Anubhuti dataset in "Fig. 4".

### A. Text Pre-processing

Pre-processing the textual content is the first step towards establishing a model for classifying the Anubhuti sentences. To pre-process the text, we make every sentence go through 2 main steps:

1. Tokenization
2. Feature Vector Representation

Tokenization is the process of splitting sentences into individual tokens or words so that Machine Learning models can work at a token-based level. Commonly used English word tokenizers are not applicable on Anubhuti dataset, hence we used the Classical Language Toolkit tokenizer for the Bengali language. Once we obtain tokenized values, we move on the very important step of representing them as a feature vector. For the baseline Machine Learning models, we used TFIDF scores of unigrams as features for each sentence. For Deep Learning Models, we used pre-trained word embeddings as feature vectors which represents every word in the input text as a dense vector of real values. FastText word embedding which is pre-trained on Bengali Wikipedia corpus. The words in our dataset are associated with a 300D (dimensional) vector by the FastText word embedding. Whenever, we come across a word not present in the word embedding, we assign a 300D vector of zeroes to it. The words in each sentence in our dataset are represented as a matrix of these vectors which is fed into our deep learning models as inputs.

TABLE III. GENRE BASED COUNTS

| Genres | Translation | Stories | Sentences |
|---|---|---|---|
| প্রেমকাহিনী | Romance | 19 | 4251 |
| গোয়েন্দা কাহিনী | Crime | 15 | 6974 |
| রহস্যগল্প | Mystery | 20 | 4720 |
| ভৌতিক গল্প | Horror | 20 | 3909 |
| কল্পবিজ্ঞান কাহিনী | Sci-Fi | 20 | 4573 |
| সামাজিক গল্প | Social Issues | 19 | 1877 |
| নারী বিষয়ক কাহিনী | Female Issues | 4 | 599 |
| অলীক কাহিনী | Fiction | 12 | 980 |
| হাস্যকৌতুক | Comedy | 16 | 1593 |
| শিশু সাহিত্য | Children's Stories | 14 | 2648 |
| Total | - | 159 | 32124 |

Fig. 4. Genre Based Counts of stories and sentences in Anubhuti

### B. Baseline Classification Models

In this section, we describe some of the baseline Machine Learning models that we used to evaluate their performance on our dataset. We trained our dataset on 4 baseline Machine Learning Models – Logistic Regression, k Nearest Neighbours, (kNN), Support Vector Machines (SVM) with Linear Kernel and Random Forest Classifier. The models were trained after choosing the best hyperparameters on a validation set. Apart from these classic Machine Learning models, we also trained a Deep Learning Model – a Convolutional Neural Network (CNN). A hyperparameter optimization on our CNN model was done by manipulating the number of layers, number of epochs, size and number of filters, batch size, number of epochs, learning rate and dropout probability. We found the Adam optimizer to perform better than the standard SGD, hence this has been used as our optimization algorithm. The results of classical Machine Learning models and our Deep Learning model for Anubhuti dataset has been calculated for the following metrics – macro average precision, recall and F1-score. We have also mentioned the overall accuracy of each of these models. We notice that some of the traditional ML models like Logistic Regression perform slightly better than the more complex architecture of the CNN. Higher accuracy of CNN can be obtained by training for more epochs on powerful hardware.

## VI. RESULTS

The results of training of the models as mentioned in the previous sub-section have been depicted in "Fig. 5". The results show the strong performances of both classic Machine Learning models as well as the Deep Learning models like the Convolutional Neural Network on our Anubhuti dataset. We found that out of all the models trained on our dataset, Logistic Regression performed the best with an accuracy of 73%. In order to fully understand the features that were most useful for these models to distinguish between the emotional classes, we provided a list of the top 5 unigram features (and their English translations) chosen by Logistic Regression in "Fig. 6". On taking a close look at the data, we see that the word choices made by the model are very relevant indicators of the emotional categories under consideration. Words like fun and happy are good indicators of the joy category while cry and dead are relevant for sorrow category. With such good feature selection, we can safely establish that Anubhuti dataset is quite capable of producing features that are trainable by Machine Learning as well as Deep Learning models.

TABLE IV. PERFORMANCE OF BASELINE ML AND DL MODELS

| Model | Precision | Recall | F1-score | Accuracy |
|---|---|---|---|---|
| Logistic Regression | 0.76 | 0.77 | 0.75 | 0.73 |
| Linear SVM | 0.75 | 0.74 | 0.69 | 0.71 |
| K Nearest Neighbors | 0.69 | 0.67 | 0.64 | 0.64 |
| Random Forest Classifier | 0.63 | 0.51 | 0.60 | 0.60 |
| CNN | 0.68 | 0.64 | 0.63 | 0.65 |

Fig. 5. Performance of Machine Learning and Deep Learning Models on Anubhuti – Precision, Recall, F1-score and Accuracy

TABLE V. FEATURE SELECTION BY LOGISTIC REGRESSION

| Emotion | Top 5 unigram features |
|---|---|
| Joy | হাসি (laughter), খুশি (joy), মজা (fun), ভালো (good), সুখী (happy) |
| Anger | রাগ (anger), চেঁচিয়ে (screamed), বিরক্তি (annoyance), প্রতিশোধ (revenge), অপমান (insult) |
| Sorrow | কান্না (cry), চোখ (eye), দুঃখ (sadness), মৃত (dead), কেঁদে (cried) |
| Suspense | ভয় (fear), ভয়ে (fear), চিৎকার (scream), রাত (night), ভূত (ghost) |
| Neutral | ঘর (house), কাজ (work), গল্প (story), জীবন (life), গান (song) |

Fig. 6. Top 5 unigram features selected by Logistic Regression Model for each emotion category

## VII. CONCLUSION AND FUTURE WORK

With this work, we have generated a dataset for emotional analysis of Bengali short stories. The results obtained from our dataset shows us that Anubhuti can be considered as an appropriate starting point for emotional analysis of Bengali sentences, stories and other classification tasks. Even though Anubhuti is created from stories, the dataset may find its application in other domains like adding emotions to audio versions of stories. Linguists may find our dataset extremely valuable in understanding the flow of emotions in Bengali short stories and try to understand the writing style of writers of these stories. In future, we plan to implement more annotations and expand our dataset for sentiment analysis and discourse modes analysis of not just short stories, but popular novels and movie scripts as well. We strongly believe that this dataset will be highly beneficial to linguists and will become a valuable resource for emotional analysis for a low resource language like Bengali.

ACKNOWLEDGMENT

We would like to thank all the writers of these short stories and the publicly available online sources from which we gathered stories for our dataset. We would also like to thank Google Colaboratory for providing powerful hardware to train our machine Learning and Deep Learning models.